\documentclass{article}

\usepackage[final, nonatbib]{neurips_2020_ml4ad}

\usepackage[utf8]{inputenc} 
\usepackage[T1]{fontenc}    
\usepackage{hyperref}       
\usepackage{url}            
\usepackage{booktabs}       
\usepackage{amsfonts}       
\usepackage{nicefrac}       
\usepackage{microtype}      

\title{Predicting times of waiting on red signals using BERT}

\author{%
  Witold Szejgis\\
  TensorCell\\
  \texttt{witoldszejgis@gmail.com} \\
  \And
  Anna Warno \\
  TensorCell\\
  Faculty of Mathematics, Informatics and Mechanics\\
  University of Warsaw\\
  \texttt{aw383513@students.mimuw.edu.pl}
  \And
  Paweł Gora \\
  TensorCell \\
  Faculty of Mathematics, Informatics and Mechanics\\
  University of Warsaw\\
  \texttt{p.gora@mimuw.edu.pl}\\
}

\begin{document}

\maketitle

\begin{abstract}
We present a method for approximating outcomes of road traffic simulations using BERT-based models, which may find applications in, e.g., optimizing traffic signal settings, especially with the presence of autonomous and connected vehicles. The experiments were conducted on a dataset generated using the Traffic Simulation Framework software runs on a realistic road network. The BERT-based models were compared with $4$ other types of machine learning models (LightGBM, fully connected neural networks and $2$ types of graph neural networks) and gave the best results in terms of all the considered metrics.
\end{abstract}

\section{Introduction}

In this article, we focus on one of the most important aspects of traffic management - traffic signal control. We present a new algorithm based on BERT for estimating the times of waiting on red signals depending on the traffic signal settings. 

In general, finding the optimal settings of traffic signals is an NP-hard problem even for relatively simple mathematical models of traffic (in reality, when traffic conditions may dynamically change, even defining what the optimality of signal settings actually mean can be difficult) \cite{nphard}. However, assuming that we can accurately predict times of waiting on red signals for different traffic signal settings, we can explore a large space of possible solutions using metaheuristics (e.g., genetic algorithms) in order to find heuristically optimal settings. Usually, the quality of settings can be evaluated using traffic simulations \cite{tsf}, but in the case of large road networks and evaluating traffic for long time periods, this method can be too time-demanding \cite{pardel}. One of the recent and interesting approaches to solve this issue is to use surrogate models based on machine learning (e.g., neural networks or LightGBM) to approximate the outcomes of traffic simulations \cite{GORA1, GORA2, GORA3, GORA4}. Such methods can be very efficient, as they return the results of evaluations a few orders of magnitude faster than in the case of computer simulations while preserving a good accuracy of approximations \cite{GORA1, GORA2, GORA3, GORA4}. For example, the recent works showed that it may be possible to approximate the outcomes of traffic simulations where the input is a vector representing signal settings and the output computed by the simulations is the total time of waiting on red signals in a given urban area and in a given time period \cite{GORA1, GORA2, GORA3, GORA4}. However, these works assumed that durations of green and red signal phases are constant (e.g., lasting $58$ and $62$ seconds, respectively \cite{GORA1}) and the only possible way of controlling traffic signal settings is by modifying traffic signal offsets (time in seconds from the beginning of a simulation to the first switch from the red signal state to the green signal state for a representative signal from a given intersection). In this paper, we extend this approach in order to consider not only various offsets but also various durations of green signal phases. For this task, we developed a new method for approximating traffic simulation outcomes based on the usage of the pretrained BERT model \cite{BERT}. 

BERT is a transformer neural network using attention mechanism which has been successful in natural language processing \cite{BERT}. It has been  pretrained on Wikipedia and BookCorpus and can be used in a number of downstream tasks like sentence and token classification, or question answering, and in many cases it achieved state of the art results \cite{BERT}. However, in this paper, we trained BERT to approximate outcomes of traffic simulations (the total time of waiting on red signals) for a given traffic signal setting. To the best of our knowledge, this is the first usage of BERT in such a task and probably even the first in the traffic prediction and traffic analysis domains.

In this paper, we evaluate 2 BERT-based models and compare them with other machine learning models which gave good results in the previous research works \cite{GORA1, GORA2, GORA3, GORA4} (feed forward neural networks, graph neural networks, graph convolutional neural networks, LightGBM), which are summarized in Section \ref{sec:related}. We ran a series of experiments on a large dataset generated using traffic simulations on a realistic road network. The results presented in Section \ref{sec:results} proved that the BERT-based models outperform other machine learning models. Therefore, BERT-based models can be later used as metamodels in experiments with traffic signal settings optimization using metaheuristics which is one of the goals of our future research.

In general, the presented method can be used for traffic signal control in any conditions, even without the presence of autonomous vehicles. However, it can be especially beneficial in the era of connected and autonomous vehicles when mathematical traffic models used in simulations can be more consistent with real-world traffic thanks to the knowledge about the algorithms controlling vehicles as well as the additional data gathered from the vehicles (e.g., planned routes).


\section{Related works}\label{sec:related}
Recently, a number of works using transformer neural networks with the attention mechanism in data outside of the NLP scope have been published. In computer vision, they obtained state of the art results in object detection and classification \cite{DETR}. In the context of graph data, both transformers and attention mechanism \cite{GAT} have been successfully applied in a number of different datasets \cite{GAT}, \cite{MAT}, \cite{image_trans}. BERT network has been successfully applied for images \cite{image_bert}, audio \cite{speech_bert} and graph data \cite{graph_bert}.

On the other hand, several research papers have been published in the domain of approximating traffic simulation outcomes using machine learning models such as fully connected neural networks or LightGBM \cite{GORA1, GORA2, GORA3, GORA4}. These methods assumed that durations of green signal phases are constant in both directions, so that offsets are the only modifiable parameter. However, to the best of our knowledge, there were no previous applications of BERT (or even transformer neural networks with the attention mechanism) to this task, or as surrogate models, in general.

\section{Experiments}\label{sec:results}

\subsection{Setup}
In order to conduct experiments and train BERT-based models and other machine learning models, we generated a dataset using the Traffic Simulation Framework software \cite{tsf}, which evaluated $1470972$ different signal settings on the Stara Ochota district in Warsaw with $21$ intersections with traffic signals (the road network data were acquired from the OpenStreetMap service \cite{osm}). Each evaluation was a 10-minute long simulation of a realistic traffic with 42000 cars. Each setting consisted of $63$ integer values, $3$ values per intersection: durations of green signal phases in $2$ directions (values from the set $\{20, 21,\ldots , 80\}$) and an offset (with values from the set $\{0, 1, 2, \ldots, max\}$, where $max$ depends on the sum of durations of green signal phases in both directions). The dataset is publicly available to facilitate the future research on that topic \cite{dataset}.

Then, we trained BERT-based models (Section \ref{sec:bert}) as well as $4$ types of machine learning models for comparison (Section \ref{sec:others}): LightGBM, fully connected neural networks (FCNN), Graph Convolutional Networks (GCN) and Graph Neural Networks (GNN). For every model, the input was standardized \cite{sklearn}. Also, in the case of GCN and GNN the input had a size $3x21$, in the case of LightGBM and FCNN: $63$ ($3$ values for $21$ intersections). For BERT, we added special separators after each triple of values (for each intersection) which added $20$ values and made an input of the size $83$. In most cases, the output (the total time of waiting on red signals, in seconds) was not normalized (normalization gave worse performance), the only exception was BERT's one-stage method.

About 80\% of the dataset ($1176776$ elements) was used for training, $147098$ of elements were used for validation and the remaining $147098$ elements were used as a test set. The results of training were evaluated on a test set and compared using $3$ metrics: MAPE, MAXPE (maximum percentage error), MAXPE99 (maximum percentage error among best 99\% results - $99$-th percentile).

\subsection{Experiments with BERT}\label{sec:bert}
 
Standard BERT language model (BERT-base-uncased) from the Hugging Face library \cite{hug} was taken as a starting point of the training. Due to BERT's limitations \cite{BERT}, an offset of 200 was added to all node features, so they don't overlap with range reserved for special tokens like <CLS> or <SEP>. The triples of features of each node were  separated by <SEP> token.
 
We tested $2$ types of methods for training BERT: \textbf{Two-step method} and \textbf{One-step method}. The two-step method uses a typical way of using BERT network in regression problems. In the first step, classification is performed on bucketized outputs. To obtain class categories, a continuous output is first normalized using min-max scaler and then divided into $15$ equally separated buckets. Using these labels, BERT network is trained in a standard manner for BERT sentence classification. In the second step, regression is performed on the BERT embeddings generated for each training input by model from step 1. The regression network is a fully connected neural network with dropout. The initial classification training was performed for $15$ bucketized classes. In the classification phase, optimization was run for $15$ epochs using Adam optimizer, learning rate set to $2e-5$, weight decay of $0.01$ and batch size set to $100$. The regression network was $(768,512,1)$ a fully connected network with dropout equal to $0.05$. In the regression phase, learning rate was set to $0.01$ and the training was performed for $12$ epochs.

In the One-step method, BERT is built into a fully connected neural network, which in each step of training generates embedding and performs a backpropagation with updating of BERT weights. A regression part of the network has analogous architecture as in the two-step method with a batch size set to $64$, a learning rate of $5e-5$, and a training time of 12 epochs. 

The results for both models are presented in Table \ref{tab:results}.

\subsection{Experiments with other models}\label{sec:others}
Experiments with $4$ other models have been conducted. LightGBM \cite{ke} and FCNN were used as benchmarks for regression problem as well as GCN \cite{kipf} and GNN \cite{mlinpl} state-of-the-art architectures suited for graph structures which road crossings form.

\subsubsection{LightGBM}
Implementation from LightGBM python package was used \cite{ke}. Hyperparameters were found using Hyperopt \cite{hyperopt} (based on Bayesian approach) for $300$ iterations with k-fold cross validation for $k=5$, mean squared error was used as a loss function. The following parameters were investigated: \texttt{learning\_rate} - uniform (0.001, 0.8), \texttt{max\_depth} - integer (3, 20), \texttt{min\_child\_weight} - integer (1, 20), \texttt{colsample\_by\_tree} - uniform (0.3, 0.8), \texttt{subsample} - uniform (0.8, 1),
 \texttt{n\_estimators} - \{100, 250, 500\}. Results for the model with optimal parameters (\texttt{colsample\_bytree} = 0.797, \texttt{learning\_rate} = 0.164, \texttt{max\_depth}= 9, \texttt{min\_child\_weight} = 1,\texttt{n\_estimators} =  500, \texttt{subsample} = 0.814) are presented in Table \ref{tab:results}.


\subsubsection{FCNN} Neural networks were trained with three different activations functions (relu, leakyrelu, tanh). Architectures with non-increasing sequences of neurons in hidden layers from the set \{256, 128, 64, 48, 32, 16, 8, 4\} with a maximum length of 9 were tested. Networks were trained with batch normalization. MSE was used as a loss function, Adam optmizer with ReduceLROnPlateau (factor = 0.2, patience = 2) and starting learning rate = 0.05, batch size = 2048 was chosen as an optimizer. The early stopping mechanism was applied, so the number of training epochs was different for different activation functions and did not exceed 100. Results for the best model (FCNN - leakyrelu, (63, 256, 128, 64, 48, 32, 16,  8, 1), neurons in hidden layers) are presented in Table \ref{tab:results}.

 \subsubsection{Graph Convolutional Network} \label{GCN}
 GCN architecture was based on graph convolutional layers \cite{kipf} with normalized adjacency matrix followed by dense linear layers. Networks with 1-4 graph convolution layers and non-increasing sequences of neurons in dense linear layers from the set \{ 64, 48, 32, 16, 8, 4 \} with maximum length of 6, relu or leakyrelu activation function were tested. MSE was used as a loss function, Adam optmizer with ReduceLROnPlateau (factor = 0.2, patience = 2) and starting learning rate = 0.05, batch size = 2048 was chosen as an optimizer. The number of training epochs was different for different activation functions (the early stopping mechanism was implemented) but did not exceed $100$. Results for the best model (GCN with 4 graph convolutionals layers and (21, 128, 48, 32) neurons in hidden layers with activation function leakyrelu) are presented in Table \ref{tab:results}.

 \subsubsection{GNN} 
 This architecture was a graph neural network inspired by \cite{mlinpl}. It had sparse layers where connections from a neuron in one layer to a neuron in the next layer is only present if the corresponding vertices are neighbors in the road network graph. The tanh activation function was used for the neurons corresponding to road network intersections (this activation function was proven to be superior in similar tasks \cite{GORA4}), followed by dense linear layers with leakyrelu or relu activation function. Architectures with 1-4 such sparse layers, 1-4 channels and the same setup as for GCN were tested. The results are presented in Table \ref{tab:results}.

\begin{table}[h]
  \caption{Results for BERT based methods}
  \label{tab:results}
  \centering
  \begin{tabular}{lllll}
    \toprule
    Model & RMSE & MAPE & MAXPE & MAXPE99  \\
    \midrule
    BERT one-step     & 1305  & 1.99\% & 17.79\% & 6.64\% \\
    BERT two-step     & 1308  & 1.99\% & 22.35\% & 10.83\%  \\
    LightGBM & 2176 & 3.4\% & 27.1\%  & 11.7\%  \\
    FCNN     & 2369  & 3.7\% & 27.1\% & 11.7\% \\
    GCN     & 	5120  & 7.78\% & 44.2\% & 25.14\%  \\
    GNN     & 	4515  & 6.86\% & 41.65\% & 22.91\%  \\
    \bottomrule
  \end{tabular}
\end{table}

\section{Conclusions}\label{sec:conclusions}
We presented a method for approximating outcomes of traffic simulations for different settings of traffic signals using pretrained BERT-based models. Such models outperformed other machine learning methods (which were proven successful in similar tasks, but without modifiable green signal phase durations) in terms of mean percentage error and maximal errors (for the whole test set and the top 99\% of the test set). The one-step method outperforms other models.

The achieved results have the potential of being used to build the next generation of traffic signal control systems that may be especially successful in the era with connected and autonomous vehicles. The knowledge about vehicle control algorithms and availability of traffic data may help in building traffic simulation models more consistent with actual traffic.

As for future work, we are planning to test more BERT-based (and transformer-based) architectures. In addition, we will run experiments with metaheuristics (e.g., genetic algorithms) searching space of traffic signal settings in order to find the best signal control solutions, in which the setting's evaluation will be done using the trained metamodels (a direct application to the traffic signal control task). Also, we are going to work on scalability of the presented method and run experiments for other urban districts and different road network topologies.

\end{document}